\documentclass{article}
\usepackage{nips12submit_e,bm,amsmath,amsfonts,tikz,amssymb,amsthm,array,todonotes,times}
\usepackage{hyperref}
\usepackage[english]{babel}

\newtheorem{theorem}{Theorem}

\newcommand{\head}[2]{\multicolumn{1}{>{\centering\arraybackslash}p{#1}}{\textbf{#2}}}
\newcommand{\augx}{\bm \Phi_{\bm P(\bm X)}^{k,s}}
\newcommand{\augy}{\bm \Phi_{\bm P(\bm Y)}^{k,s}}

\definecolor{MPG}{RGB}{000,125,122}
\newcounter{PHcomment} 

\nipsfinalcopy

\title{The Randomized Dependence Coefficient}

\author{
  \textbf{David Lopez-Paz, Philipp Hennig, Bernhard Sch\"olkopf}\\
  Max Planck Institute for Intelligent Systems\\
  Spemannstra{\ss}e 38, T\"ubingen, Germany\\
  \texttt{\{dlopez,phennig,bs\}@tue.mpg.de}
} 

\begin{document} 
 
\maketitle 
 
\begin{abstract}
  We introduce the Randomized Dependence Coefficient (RDC), a measure of
  non-linear dependence between random variables of arbitrary dimension based
  on the Hirschfeld-Gebelein-R\'enyi Maximum Correlation Coefficient. RDC is
  defined in terms of correlation of random non-linear copula
  projections; it is invariant with respect to marginal distribution
  transformations, has low computational cost and is easy to implement: just
  five lines of R code, included at the end of the paper.
\end{abstract}

\section{Introduction}\label{sec:intro}
Measuring statistical dependence between random variables is a fundamental
problem in statistics. Commonly used measures of dependence, Pearson's rho,
Spearman's rank or Kendall's tau are computationally efficient and
theoretically well understood, but consider only a limited class of association
patterns, like linear or monotonically increasing functions. The development of
non-linear dependence measures is challenging because of the radically larger
amount of possible association patterns.

Despite these difficulties, many non-linear statistical dependence measures
have been developed recently. Examples include the Alternating Conditional
Expectations or \emph{backfitting algorithm} (ACE) \cite{Breiman85,Hastie86},
Kernel Canonical Correlation Analysis (KCCA) \cite{Bach02}, (Copula) Maximum
Mean Discrepancy (MMD, CMMD in their HSIC formulations)
\cite{Gretton05,Gretton12,Poczos12}, Distance or Brownian Correlation (dCor)
\cite{Szekely07,Szekely10} and the Maximal Information Coefficient (MIC)
\cite{Reshef11}. However, these methods exhibit high computational demands (at
least quadratic costs in the number of samples for KCCA, MMD, CMMD, dCor or
MIC), are limited to measuring dependencies between scalar random variables
(ACE, MIC), show poor performance under the existence of additive noise (MIC)
or can be difficult to implement (ACE, MIC).

This paper develops the \emph{Randomized Dependence Coefficient} (RDC), an
estimator of the Hirschfeld-Gebelein-R\'enyi Maximum Correlation Coefficient
(HGR) addressing the issues listed above. RDC defines dependence between two
random variables as the largest canonical correlation between random non-linear
projections of their respective empirical copula-transformations. RDC is
invariant to monotonically increasing transformations, operates on random
variables of arbitrary dimension, and has computational cost of $O(n\log n )$
with respect to the sample size.  Moreover, it is easy to implement: just five
lines of R code, included in Appendix \ref{sec:code}.

The following Section reviews the classic work of Alfr\'ed R\'enyi
\cite{Renyi59}, who proposed seven desirable fundamental properties of
dependence measures, proved to be satisfied by the
Hirschfeld-Gebelein-R\'enyi's Maximum Correlation Coefficient (HGR). Section
\ref{sec:rdc} introduces the Randomized Dependence Coefficient as an estimator
designed in the spirit of HGR, since HGR itself is computationally intractable.
Properties of RDC and its relationship to other non-linear dependence measures
are analysed in Section \ref{sec:rdc_prop}. Section \ref{sec:exps} validates
the empirical performance of RDC on a series of numerical experiments on both
synthetic and real-world data.

\section{Hirschfeld-Gebelein-R\'enyi's Maximum Correlation
Coefficient}\label{sec:renyi}
In 1959 \cite{Renyi59}, Alfr\'ed R\'enyi argued that a measure of dependence
$\rho^* : \mathcal{X} \times \mathcal{Y} \rightarrow [0,1]$ between random
variables $X\in\mathcal{X}$ and $Y\in\mathcal{Y}$ should satisfy seven
fundamental properties:
\begin{enumerate}
  \item $\rho^*(X,Y)$ is defined for any pair of non-constant random variables
  $X$ and $Y$.
  \item $\rho^*(X,Y) = \rho^*(Y,X)$
  \item $0 \leq \rho^*(X,Y) \leq 1$
  \item $\rho^*(X,Y) = 0$ iff $X$ and $Y$ are statistically independent.
  \item For bijective  Borel-measurable functions $f,g : \mathbb{R}
  \rightarrow \mathbb{R}$, $\rho^*(X,Y) = \rho^*(f(X),g(Y))$.
  \item $\rho^*(X,Y) = 1$ if for Borel-measurable functions $f$ or $g$, $Y =
  f(X)$ or $X = g(Y)$.
  \item If $(X,Y) \sim \mathcal{N}(\bm \mu, \bm \Sigma)$, then $\rho^*(X,Y) =
  |\rho(X,Y)|$, where $\rho$ is the correlation coefficient.
\end{enumerate}
R\'enyi also showed the \emph{Hirschfeld-Gebelein-R\'enyi Maximum
  Correlation Coefficient} (HGR) \cite{Gebelein41,Renyi59} to satisfy
all these properties. HGR was defined by Gebelein in 1941
\cite{Gebelein41} as the supremum of Pearson's correlation coefficient
$\rho$ over all Borel-measurable functions $f,g$ of finite variance:
\begin{equation}\label{eq:hgr}
  \text{hgr}(X,Y) = \sup_{f,g} \rho(f(X),g(Y)),
\end{equation}
Since the supremum in \eqref{eq:hgr} is over an infinite-dimensional
space, HGR is not computable. It is an abstract concept, not a
practical dependence measure. In the following we propose a scalable
estimator with the same structure as HGR: the Randomized Dependence
Coefficient.

\section{Randomized Dependence Coefficient} \label{sec:rdc} 
\begin{figure}[h!]
  \includegraphics[width=\textwidth]{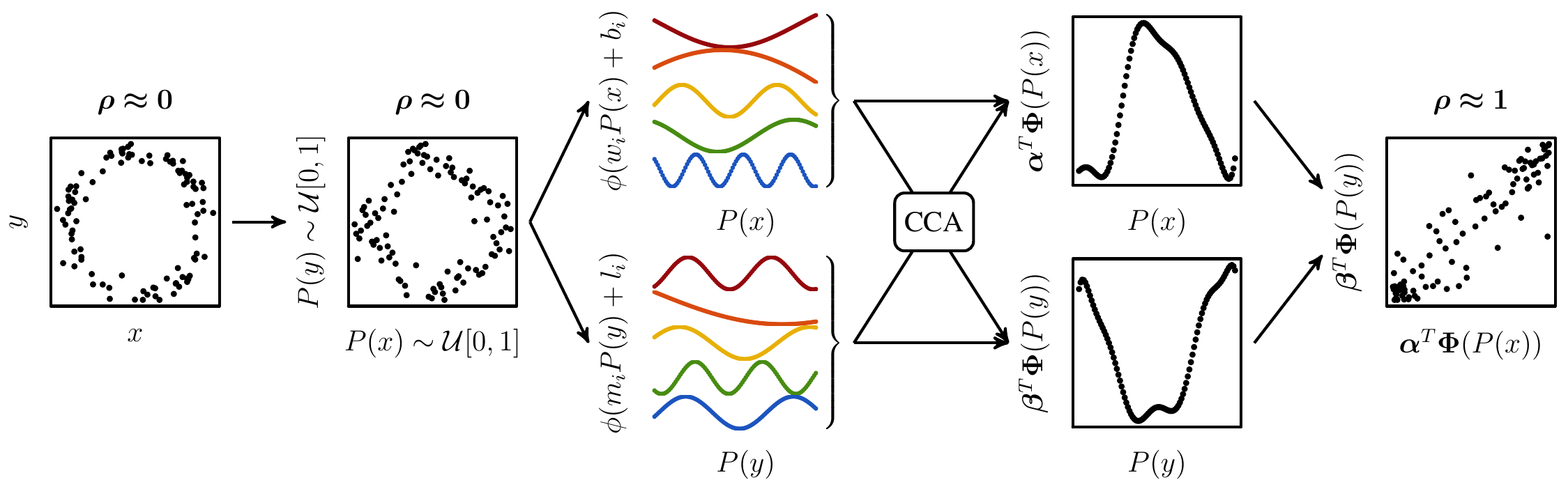}
  \caption{RDC computation for a simple set of samples
  $\{(x_i,y_i)\}_{i=1}^{100}$ drawn from a noisy circular pattern: The samples
  are used to estimate the copula, then mapped with randomly drawn non-linear
  functions. The RDC is the largest canonical correlation between these
  non-linear projections.}
  \label{fig:rdcsteps}
\end{figure}

The \emph{Randomized Dependence Coefficient} (RDC) measures the dependence
between random samples $\bm X \in \mathbb{R}^{p\times n}$ and $\bm Y \in
\mathbb{R}^{q\times n}$ as the largest canonical correlation between $k$
randomly chosen non-linear projections of their copula transformations. Before
Section~\ref{sec:formal-definition-or} defines this concept formally, we
describe the three necessary steps to construct the RDC statistic:
copula-transformation of each of the two random samples
(Section~\ref{sec:estim-copula-transf}), projection of the copulas through $k$
randomly chosen non-linear maps (Section~\ref{sec:gener-rand-non}) and
computation of the largest canonical correlation between the two sets of
non-linear random projections (Section~\ref{sec:comp-canon-corr}). Figure
\ref{fig:rdcsteps} offers a sketch of this process.

\subsection{Estimation of Copula-Transformations}
\label{sec:estim-copula-transf}
To achieve invariance with respect to transformations on marginal distributions
(such as shifts or rescalings), we operate on the \emph{empirical copula
transformation} of the data \cite{Nelsen06,Poczos12}. Consider a random vector
$\bm X = (X_1, \ldots, X_d)$ with continuous marginal cumulative distribution
functions (cdfs) $P_i$, $1 \leq i \leq d$. Then the vector $\bm U =
(U_1,\ldots,U_d) := \bm P(\bm X) = (P_1(X_1),\ldots,P_d(X_d))$, known as the
\emph{copula transformation}, has uniform marginals:
\begin{theorem}
  (Probability Integral Transform \cite{Nelsen06}) For a random variable $X$
  with cdf $P$, the random variable $U := P(X)$ is uniformly distributed on
  $[0,1]$.
\end{theorem}
The random variables $U_1, \ldots, U_d$ are known as the observation ranks of
$X_1, \ldots, X_d$. Crucially, $\bm U$ preserves the dependence structure of
the original random vector $\bm X$, but ignores each of its $d$ marginal forms
\cite{Nelsen06}.  The joint distribution of $\bm U$ is known as the copula of
$\bm X$:
\begin{theorem}
  (Sklar \cite{Sklar59}) Let the random vector $\bm X = (X_1, \ldots, X_d)$
  have continuous marginal cdfs $P_i$, $1 \leq i \leq d$. Then, the joint
  cumulative distribution of $\bm X$ is uniquely expressed as:
\begin{equation}
  P(X_1, \ldots, X_d) = C(P_1(X_1), \ldots, P_d(X_d)),
\end{equation}
where the distribution $C$ is known as the copula of $\bm X$.
\end{theorem}
A practical estimator of the univariate cdfs $P_1, \ldots, P_d$ is the
\emph{empirical cdf}:
\begin{equation}
  P_n(x) := \frac{1}{n} \sum_{i=1}^n \mathbb{I}(X_i \leq x),
\end{equation}
which gives rise to the \emph{empirical copula transformations} of a
multivariate sample:
\begin{equation}
\bm P_n(\bm x) = [{P}_{n,1}(x_1), \ldots, {P}_{n,d}(x_d)].
\end{equation}
The Massart-Dvoretzky-Kiefer-Wolfowitz inequality \cite{Massart90} can be used
to show that empirical copula transformations converge fast to the true
transformation as the sample size increases:
\begin{theorem}
  (Convergence of the empirical copula, \cite[Lemma 7]{Poczos12}) Let $\bm X_1,
  \ldots, \bm X_n$ be an i.i.d. sample from a probability distribution over
  $\mathbb{R}^d$ with marginal cdf's $P_1, \ldots, P_d$. Let $\bm P(\bm X)$ be
  the copula transformation and $\bm P_n(\bm X)$ the empirical copula
  transformation. Then, for any $\epsilon > 0$:
    \begin{equation}
      \Pr \left[ \sup_{\bm x \in \mathbb{R}^d} \| \bm P(\bm x) - \bm P_n(\bm x)
      \|_2 > \epsilon \right] \leq 2 d \exp \left( -\frac{2 m
      \epsilon^2}{d}\right).
    \end{equation}
\end{theorem}
Computing $\bm P_n(\bm X)$ involves sorting the marginals of $\bm X \in
\mathbb{R}^{d\times n}$, thus $O(d n \log (n))$ operations.

\subsection{Generation of Random Non-Linear Projections}
\label{sec:gener-rand-non}
The second step of the RDC computation is to augment the empirical copula
transformations with non-linear projections, so that linear methods can
subsequently be used to capture non-linear dependencies on the original data.
This is a classic idea also used in other areas, particularly in regression. In
an elegant result, Rahimi and Brecht \cite{Rahimi08} proved that linear
regression on random, non-linear projections of the original feature space can
generate high-performance regressors:
\begin{theorem}\label{thm:rahimi}
(Rahimi-Brecht) 
Let $p$ be a distribution on $\Omega$ and $|\phi(\bm x; \bm w)| \leq 1$.
Let $\mathcal{F} = \left\lbrace \left. f(\bm x) = \int_\Omega
    \alpha(\bm w) \phi(\bm x; \bm w) \mathrm{d}\bm w \right|
  |\alpha(\bm w)| \leq C p(\bm w)\right\rbrace$.
Draw $\bm w_1, \ldots, \bm w_k$ iid from $p$.
Further let $\delta > 0$, and $c$ be some $L$-Lipschitz loss function,
and consider data $\{\bm x_i, y_i\}_{i=1}^n$ drawn iid from some
arbitrary $P(\bm X,Y)$. The $\alpha_1, \ldots, \alpha_k$ for which
${f}_k(\bm x) = \sum_{i=1}^k \alpha_i \phi(\bm x; \bm w_i)$ minimizes
the empirical risk $c(f_k(\bm x),y)$ has a distance from the $c$-optimal
estimator in $\mathcal{F}$ bounded by
\begin{equation}
 \mathbb{E}_P[c({f}_k(\bm x),y)] - \min_{f\in \mathcal{F}}
 \mathbb{E}_P[c(f(\bm x),y)] \leq O\left(\left( \frac{1}{\sqrt{n}} +
 \frac{1}{\sqrt{k}} \right) LC\sqrt{\log \frac{1}{\delta}}\right)
\end{equation}
with probability at least $1-2\delta$.
\end{theorem}
Intuitively, Theorem \ref{thm:rahimi} states that randomly selecting $\bm w_i$
in $\sum_{i=1}^k \alpha_i \phi(\bm x; \bm w_i)$ instead of optimising them
causes only bounded error.

The choice of the non-linearities $\phi : \mathbb{R} \rightarrow \mathbb{R}$ is
the main, unavoidable assumption in RDC. This choice is a well-known problem
common to all non-linear regression methods and has been studied extensively in
the theory of regression as the selection of reproducing kernel Hilbert space
\cite[\textsection3.13]{learningwithkernels}. The choice of the family (space)
of features, and of probability distributions over it, is unlimited. The only
way to favour one such family and distribution over another is to use prior
assumptions about which kind of distributions the method will typically have to
analyse. 

Features popular in parts of the literature are sigmoids, parabolas, radial
basis functions, complex sinusoids, sines or cosines. In our experiments, we
will use sine and cosine projections, $\phi(\bm w^T \bm x +b) := (\cos(\bm w^T
\bm x + b), \sin(\bm w^T \bm x + b))$.  Arguments favouring this choice are
that shift-invariant kernels are approximated with these features when using
the appropriate random parameter sampling distribution
\cite{Rahimi08},\cite[p.~208]{gihman4s:_theor_stoch_proces}
\cite[p.~24]{stein99:_inter_spatial_data}, and that functions with absolutely
integrable Fourier transforms are approximated with $L_2$ error below
$O(1/\sqrt{k})$ by $k$ of these features \cite{Jones92}.

Let the random parameters $\bm w_i \sim \mathcal{N}(\bm 0, s\bm I)$, $b_i \sim
\mathcal{U}[-\pi,\pi]$. Choosing $\bm w_i$ to be Normal is analogous to the use
of the Gaussian kernel for MMD, CMMD or KCCA \cite{Rahimi08}.
Tuning $s$ is analogous to selecting the kernel width, that is, to regularize the
non-linearity of the random projections.

Given a data collection $\bm X = (\bm x_1, \ldots, \bm x_n)$, we will
denote by
\begin{equation}
  \bm \Phi(\bm X; k,s) := 
  \left(
  \begin{array}{ccc}
  \phi(\bm w_1^T \bm x_1+b_1) & \cdots & \phi(\bm w_k^T \bm x_1+b_k)\\
  \vdots & \vdots & \vdots \\
  \phi(\bm w_1^T \bm x_n+b_1) & \cdots & \phi(\bm w_k^T \bm x_n +b_k)
  \end{array}
  \right)^T
\end{equation}
the $k-$th order random non-linear projection from $\bm X \in \mathbb{R}^{d
\times n}$ to $\bm \Phi^{k,s}_{\bm X} := \bm \Phi(\bm X; k, s) \in
\mathbb{R}^{2k \times n}$. The computational complexity of computing $\bm
\Phi^{k,s}_{\bm X}$ using naive matrix multiplications is $O(kdn)$. However,
recent techniques \cite{Le13} allow computing these feature expansions within a
computational cost of $O(k\log(d)n)$ using $O(k)$ storage.

\subsection{Computation of Canonical Correlations} \label{sec:comp-canon-corr}
The final step of RDC is to compute the linear combinations of the augmented
empirical copula transformations that have maximal correlation. Canonical
Correlation Analysis (CCA, \cite{Haerdle07}) is the calculation of pairs of
basis vectors $(\bm \alpha, \bm \beta)$ such that the projections $\bm \alpha^T
\bm X$ and $\bm \beta^T \bm Y$ of two random samples $\bm X \in
\mathbb{R}^{p\times n}$ and $\bm Y \in \mathbb{R}^{q\times n}$ are maximally
correlated. The correlations between the projected (or canonical) random
samples are referred to as canonical correlations. There exist up to
$\max(\text{rank}(\bm X), \text{rank}(\bm Y))$ of them. Canonical correlations
$\rho^2$ are the solutions to the eigenproblem:
\begin{align}
  \left(
    \begin{array}{cc}
    \bm 0 & \bm C_{xx}^{-1}\bm C_{xy}\\
    \bm C_{yy}^{-1}\bm C_{yx} & \bm 0
    \end{array}
  \right)
  \left(
    \begin{array}{c}
    \bm \alpha\\
    \bm \beta 
    \end{array}
  \right)
  =
  \rho^2
  \left(
    \begin{array}{c}
    \bm \alpha\\
    \bm \beta 
    \end{array}
  \right)
  ,\label{eq:eigenset}
\end{align}
where $\bm C_{xy} = \text{cov}(\bm X, \bm Y)$ and the matrices $\bm C_{xx}$ and
$\bm C_{yy}$ are assumed to be invertible. Therefore, the largest canonical
correlation $\rho_1$ between $\bm X$ and $\bm Y$ is the supremum of the
correlation coefficients over their linear projections, that is:
$  \rho_1(\bm X, \bm Y) = \sup_{\bm \alpha, \bm \beta} \rho(\bm \alpha^T \bm X,
  \bm \beta^T \bm Y).
$

When $p,q \ll n$, the cost of CCA is dominated by the estimation of the matrices $\bm
C_{xx}$, $\bm C_{yy}$ and $\bm C_{xy}$, hence being $O((p+q)^2n)$ for two
random variables of dimensions $p$ and $q$, respectively.

\subsection{Formal Definition or RDC}
\label{sec:formal-definition-or}
Given the random samples $\bm X \in \mathbb{R}^{p\times n}$ and $\bm Y \in
\mathbb{R}^{q\times n}$ and the parameters $k \in \mathbb{N}_+$ and $s \in
\mathbb{R}_+$, the Randomized Dependence Coefficient between $\bm X$ and $\bm
Y$ is defined as:
\begin{equation}\label{eq:rdc}
  \text{rdc}(\bm X, \bm Y; k,s) :=
  \sup_{\bm \alpha, \bm \beta}\rho\left(\bm \alpha^T \augx, \bm \beta^T
  \augy\right).
\end{equation}

\section{Properties of RDC}\label{sec:rdc_prop}
\paragraph{Computational complexity:} In the typical setup (very large $n$,
large $p$ and $q$, small $k$) the computational complexity of RDC is dominated
by the calculation of the copula-transformations. Hence, we achieve a cost in
terms of the sample size of $O((p+q) n \log n + kn\log(pq) + k^2n) \approx O(n
\log n)$. 

\paragraph{Ease of implementation:} An implementation of RDC in R is included
in the Appendix \ref{sec:code}.

\paragraph{Relationship to the HGR coefficient:} \label{sec:relat-hgr}
It is tempting to wonder whether RDC is a consistent, or even an efficient
estimator of the HGR coefficient. However, a simple experiment shows
that it is not desirable to approximate HGR exactly on finite datasets:
Consider $p(X,Y)=\mathcal{N}(x;0,1)\mathcal{N}(y;0,1)$ which is independent,
thus, by both R\'enyi's 4th and 7th properties, has $\mathrm{hgr}(X,Y)=0$.
However, for finitely many $N$ samples from $p(X,Y)$, almost surely, values in
both $X$ and $Y$ are pairwise different and separated by a finite difference.
So there exist continuous (thus Borel measurable) functions $f(X)$ and $g(Y)$
mapping both $X$ and $Y$ to the sorting ranks of $Y$, i.e.
$f(x_i)=g(y_i)\;\forall (x_i,y_i)\in(\bm X,\bm Y)$. Therefore, the
finite-sample version of Equation \eqref{eq:hgr} is constant and equal to ``1'' 
for continuous random variables. Meaningful measures of dependence from finite
  samples thus must rely on some form of regularization. RDC achieves this by
  approximating the space of Borel measurable functions with the restricted
  function class $\mathcal{F}$ from Theorem \ref{thm:rahimi}:

Assume the optimal transformations $f$ and $g$ (Equation 1) to belong to the
Reproducing Kernel Hilbert Space $\mathcal{F}$ (Theorem 4), with associated
shift-invariant, positive semi-definite kernel function $k(\bm x, \bm x') =
\langle \bm \phi(\bm x), \bm \phi(\bm x')\rangle_\mathcal{F} \leq 1$. Then, with
probability greater than $1-2\delta$:
\begin{equation}
\label{eq:err_total}
\mathrm{hgr}(\bm X, \bm Y; \mathcal{F}) - 
\mathrm{rdc}(\bm X, \bm Y; k) 
= O\left(\left(\frac{\|\bm
m\|_F}{\sqrt{n}}+\frac{LC}{\sqrt{k}}\right)
 \sqrt{\log\frac{1}{\delta}}\right),\end{equation}
where $\bm m := \bm \alpha \bm \alpha^T +\bm \beta \bm \beta^T$ and $n$, $k$
denote the sample size and number of random features. The bound
(\ref{eq:err_total}) is the sum of two errors. The error $O(1/\sqrt{n})$ is due
to the convergence of CCA's largest eigenvalue in the finite sample size
regime.  This result \cite[Theorem 6]{Hardoon09} is originally obtained by
posing CCA as a least squares regression on the product space induced by the
feature map $\bm \psi(\bm x, \bm y) = [\bm \phi(\bm x)\bm \phi(\bm x)^T, \bm
\phi(\bm y) \phi(\bm y)^T, \sqrt{2}\bm \phi(\bm x) \phi(\bm y)^T]^T$.  Because of
approximating $\bm \psi$ with $k$ random features, an additional error 
$O(1/\sqrt{k})$ is introduced in the least squares regression \cite[Lemma
3]{Rahimi08}. Therefore, an equivalence between RDC and KCCA is established if
RDC uses an infinite number of sine/cosine features, the random sampling distribution is
set to the inverse Fourier transform of the shift-invariant kernel used by KCCA and
the copula-transformations are discarded.
However, when $k \geq n$ regularization
is needed to avoid spurious perfect correlations, as discussed above.

\paragraph{Relationship to other estimators:}\label{sec:comparison} Table
\ref{table:comparison} summarizes several state-of-the-art dependence measures
showing, for each measure, whether it allows for general non-linear dependence
estimation, handles multidimensional random variables, is invariant with
respect to changes in marginal distributions, returns a statistic in $[0,1]$,
satisfy R\'enyi's properties (Section \ref{sec:renyi}), and how many parameters
it requires. As parameters, we here count the kernel function for kernel
methods, the basis function and number of random features for RDC, the stopping
tolerance for ACE and the search-grid size for MIC, respectively. Finally, the
table lists computational complexities with respect to sample size.

\begin{table}[h!]
  \caption{Comparison between non-linear dependence measures.}
  \vskip 0.3 cm
\resizebox{\textwidth}{!}
{
  \begin{tabular}{lccccccl}
  \hline
  \head{1.5cm}{\textbf{Name of Coeff.}}     &
  \head{ .9cm}{\textbf{Non-Linear}}         &
  \head{1.2cm}{\textbf{Vector Inputs}}      &
  \head{1.6cm}{\textbf{Marginal Invariant}} & 
  \head{1.9cm}{\textbf{Renyi's Properties}} &
  \head{1.1cm}{Coeff. \textbf{$\in [0,1]$}} &
  \head{1cm}{\# \textbf{Par.}}         &
  \head{1cm}{\textbf{Comp. Cost}}\\\hline\hline
  Pearson's $\rho$      & $\times$     & $\times$     & $\times$     & $\times$     & $\checkmark$ & 0 & $n$        \\ \hline
  Spearman's $\rho$     & $\times$     & $\times$     & $\checkmark$ & $\times$     & $\checkmark$ & 0 & $n \log n$ \\ \hline
  Kendall's $\tau$      & $\times$     & $\times$     & $\checkmark$ & $\times$     & $\checkmark$ & 0 & $n \log n$ \\ \hline
  CCA                   & $\times$     & $\checkmark$ & $\times$     & $\times$     & $\checkmark$ & 0 & $n$     \\ \hline
  KCCA \cite{Bach02}    & $\checkmark$ & $\checkmark$ & $\times$     & $\times$     & $\checkmark$ & 1 & $n^3$      \\ \hline
  ACE  \cite{Breiman85} & $\checkmark$ & $\times$     & $\times$     & $\checkmark$ & $\checkmark$ & 1 & $n$   \\ \hline
  MIC  \cite{Reshef11}  & $\checkmark$ & $\times$     & $\times$     & $\times$     & $\checkmark$ & 1 & $2^n$      \\ \hline
  dCor \cite{Szekely07} & $\checkmark$ & $\checkmark$ & $\times$     & $\times$     & $\checkmark$ & 1 & $n^2$      \\ \hline
  MMD \cite{Gretton12}  & $\checkmark$ & $\checkmark$ & $\times$     & $\times$     & $\times$     & 1 & $n^2$      \\ \hline
  CMMD \cite{Poczos12}  & $\checkmark$ & $\checkmark$ & $\checkmark$ & $\times$     & $\times$     & 1 & $n^2$      \\ \hline
  \textbf{RDC}          & $\checkmark$ & $\checkmark$ & $\checkmark$ & $\checkmark$ & $\checkmark$ & 2 & $n \log n$ \\ \hline\hline
  \end{tabular}
}
\label{table:comparison}
\end{table}

\paragraph{Testing for independence with RDC:} Consider the hypothesis ``the
two sets of non-linear projections are mutually uncorrelated''.  Under
normality assumptions (or large sample sizes), Bartlett's approximation
\cite{Mardia79} can be used to show:
\begin{equation}
  \left(\frac{2k+3}{2}-n\right) \log \prod_{i=1}^k (1-\rho_i^2) \sim
  \chi^2_{k^2},
\end{equation}
where $\rho_1, \ldots, \rho_k$ are the canonical correlations between the two
sets of non-linear projections $\augx$ and $\augy$. Alternatively, non-parametric asymptotic
distributions can be obtained from the spectrum of the inner products of the
non-linear random projection matrices \cite[Theorem 3]{Zhang12}.

\section{Experimental Results}\label{sec:exps}
We performed numerical experiments on both synthetic and real-world data to
validate the empirical performance of RDC versus the non-linear dependence
measures listed in Table \ref{table:comparison}. In some experiments, we don't compare against to
KCCA due its prohibitive running times (see Table \ref{fig:times}).

\paragraph{Parameter selection:} The number of random features for RDC was set
to $k=10$ symmetrically for both random samples, since no significant
improvements were observed for larger values.  However, this parameter can be
set to the largest value that fits within the available computational
budget. The random sampling parameters $(s_{\bm X}, s_{\bm Y})$ were set
independently for each of the two random samples, equal to their squared
euclidean distance empirical median \cite{Gretton12}.  Competing kernel methods
make use of Gaussian RBF kernels of the form $k(\bm x, \bm x'; s_{\bm X}) =
exp(-\| \bm x- \bm x'\|^2/s_{\bm X})$ for the random variable $\bm X$ and
analogously for the random variable $\bm Y$. For the MIC statistic, the
search-grid size is set to $B(n) = n^{0.6}$, as recommended by the authors
\cite{Reshef11}. The stopping tolerance for ACE is set to $\epsilon = 0.01$,
the default value in the R package
\texttt{acepack}\footnote{\url{http://cran.r-project.org/web/packages/acepack/index.html}}.

\subsection{Synthetic Data}

\begin{figure}[t!]
\includegraphics[width=\textwidth]{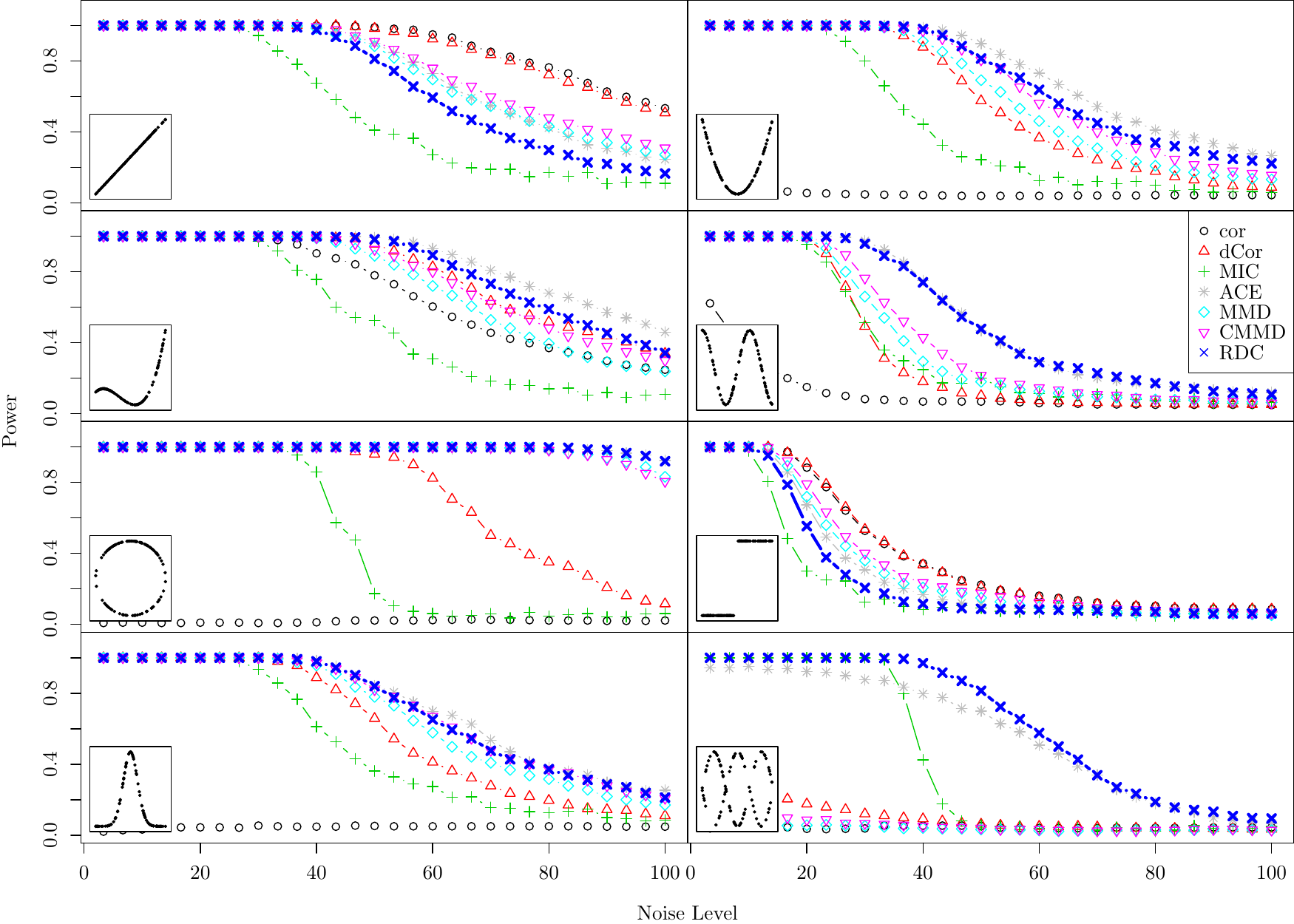}
\caption{Power of discussed measures on several bivariate association patterns
as noise increases. Insets show the noise-free form of each association
pattern.}
\label{fig:power}
\end{figure}

\paragraph{Resistance to additive noise:}
We define the \emph{power} of a dependence measure as its ability to discern
between dependent and independent samples that share equal marginal forms. In
the spirit of Simon and
Tibshirani\footnote{\url{http://www-stat.stanford.edu/~tibs/reshef/comment.pdf}},
we conducted experiments to estimate the power of RDC as a measure of
non-linear dependence.  We chose 8 bivariate association patterns, depicted inside little
boxes in Figure \ref{fig:power}. For each of the 8 association patterns, 500
repetitions of 500 samples were generated, in which the input variable was
uniformly distributed on the unit interval. Next, we regenerated the input
variable randomly, to generate independent versions of each sample with equal
marginals. Figure \ref{fig:power} shows the power for the discussed non-linear dependence
measures as the variance of some zero-mean Gaussian additive noise increases
from $1/30$ to $3$.  RDC shows worse performance in the linear association
pattern due to noise overfitting and in the step-function due to the smoothness
prior induced by the use of sine/cosine basis functions, but has good
performance in non-functional association patterns (such as the circle and the
mixture of sinusoidal waves). 

\paragraph{Running times:}
Table \ref{fig:times} summarizes running times (in seconds) for the considered
non-linear dependence measures on scalar, uniformly distributed, independent
samples of sizes $\{10^3, \ldots, 10^6\}$ when averaging over 100 runs. Single
runs above ten minutes were cancelled (empty cells in table).
In this comparison, Pearson's $\rho$, ACE, dCor and
MIC are using compiled C code, while RDC, along with MMD, CMMD and KCCA are
implemented as interpreted R code.

\begin{table}[h!]
  \caption{Average running times (in seconds) for dependence measures on
  versus sample sizes.}
  \vskip 0.3 cm
    \resizebox{\textwidth}{!}{
    \begin{tabular}{l|cccccccc}
    \hline
   \bf sample size& \bf Pearson's $\rho$ & \bf RDC      & \bf ACE       & \bf
   dCor     & \bf MMD     & \bf CMMD    & \bf MIC     & \bf KCCA\\\hline\hline
    1,000     & 0.0001  & 0.0047   &  0.0080   &  0.3417  & 0.3103  &  0.3501 & 1.0983  &  166.29 \\\hline
    10,000    & 0.0002  & 0.0557   &  0.0782   &  59.587  & 27.630  &  29.522 &  ---    &  --- \\\hline
    100,000   & 0.0071  & 0.3991   &  0.5101   &  ---     &  ---    &  ---    &  ---    &  --- \\\hline
    1,000,000 & 0.0914  & 4.6253   &  5.3830   &  ---     &  ---    &  ---    &  ---    &  --- \\\hline
    \hline
    \end{tabular}
  }
\label{fig:times}
\end{table}

\begin{figure}[h!]
\centering
\includegraphics[width=\textwidth]{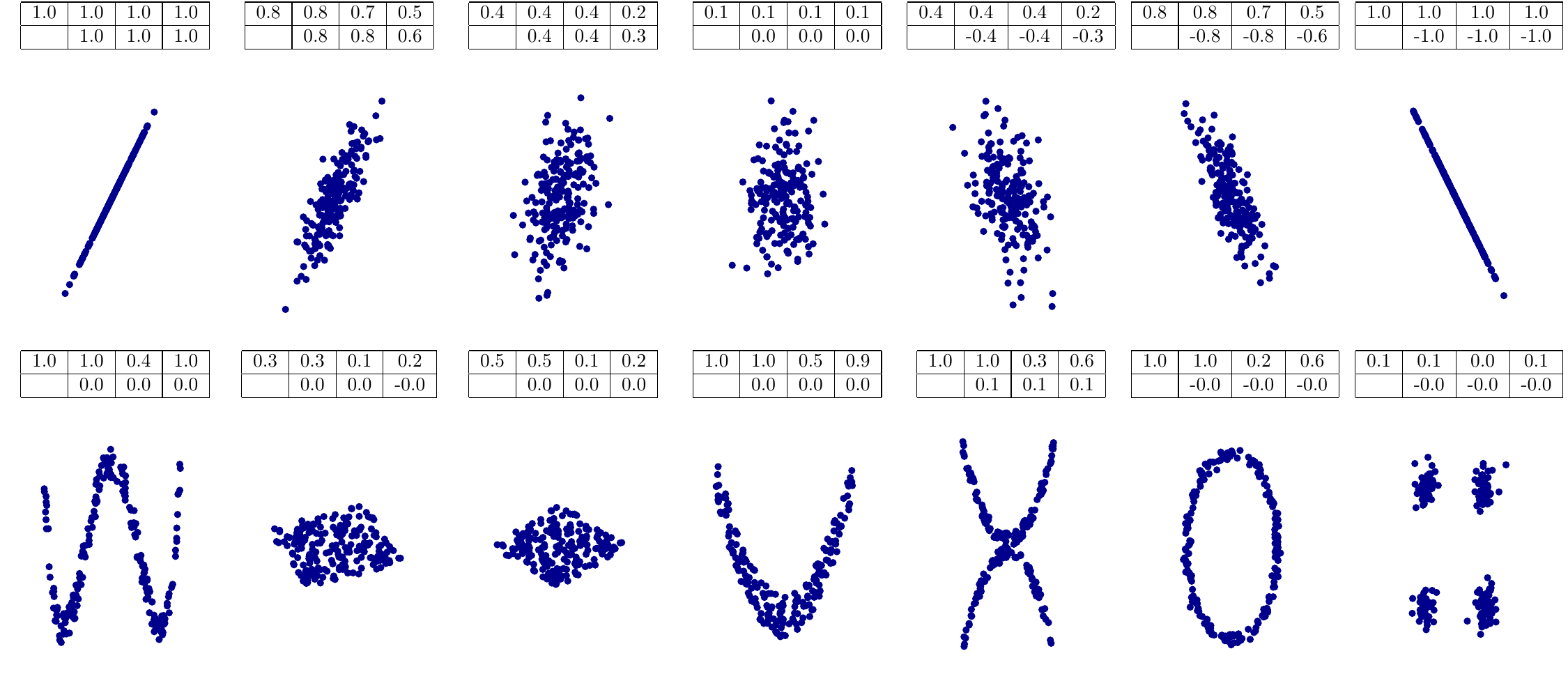}
\vskip -.5 cm
\caption{RDC, ACE, dCor, MINE, Pearson's $\rho$, Spearman's rank and Kendall's
$\tau$ estimates (numbers in tables above plots, in that order) for several
bivariate association patterns.}
\label{fig:pairs}
\end{figure}

\paragraph{Value of statistic in $[0,1]$:} Figure \ref{fig:pairs} shows RDC,
ACE, dCor, MIC, Pearson's $\rho$, Spearman's rank and Kendall's $\tau$
dependence estimates for 14 different associations of two scalar random
variables.  RDC scores values close to one on all the proposed dependent
associations, whilst scoring values close to zero for the independent
association, depicted last.  When the associations are Gaussian (first row),
RDC scores values close to the Pearson's correlation coefficient, as suggested
in the seventh property of R\'enyi (Section \ref{sec:renyi}).

\subsection{Feature Selection in Real-World Data} We performed greedy feature
selection via dependence maximization \cite{Song12} on eight real-world
datasets. More specifically, we attempted to construct the subset of features
$\mathcal{G} \subset \mathcal{X}$ that minimizes the normalized mean squared
regression error (NMSE) of a Gaussian process regressor. We do so by selecting
the feature $x^{(i)}$ maximizing dependence between the feature set
$\mathcal{G}_{i} = \{\mathcal{G}_{i-1} , x^{(i)}\}$ and the target variable $y$
at each iteration $i \in \{1, \ldots 10\}$, such that $\mathcal{G}_0 = \{
  \emptyset \}$ and $x^{(i)} \notin \mathcal{G}_{i-1}$.

We considered 12 heterogeneous datasets, obtained from the UCI dataset
repository\footnote{\url{http://www.ics.uci.edu/~mlearn}}, the Gaussian process
web site Data\footnote{\url{http://www.gaussianprocess.org/gpml/data/}} and the
Machine Learning data set repostitory\footnote{\url{http://www.mldata.org}}.
Random training/test partitions are computed to be disjoint and equal sized.

Since $\mathcal{G}$ can be multi-dimensional, we compare RDC to the non-linear
methods dCor, MMD and CMMD. Given their quadratic computational demands, dCor,
MMD and CMMD use up to $1,000$ points when measuring dependence; this
constraint only applied on the \texttt{sarcos} and \texttt{calcensus} datasets.
Results are averages of $20$ random training/test partitions.

\begin{figure}[h!]
  \includegraphics[width=\textwidth]{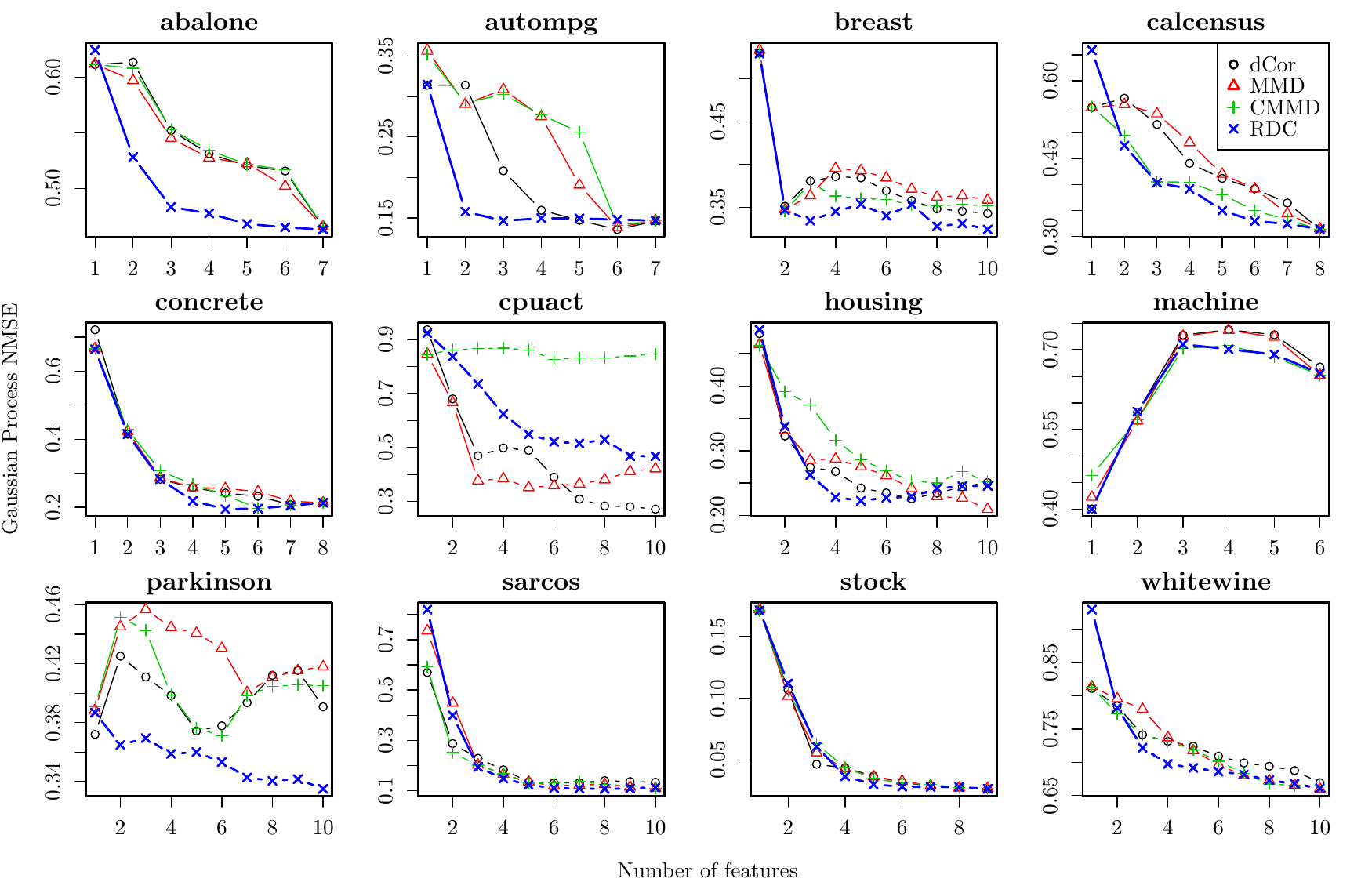}
  \caption{Feature selection experiments on real-world datasets.}
  \label{fig:featsel}
\end{figure}

Figure \ref{fig:featsel} summarizes the results for all datasets and algorithms
as the number of selected features increases. RDC performs best in most
datasets, with much lower running time than its
contenders.

\section{Conclusion}
\label{sec:conclusion}

We have presented the randomized dependence coefficient, a lightweight
non-linear measure of dependence between multivariate random samples.
Constructed as a finite-dimensional estimator in the spirit of the
Hirschfeld-Gebelein-R\'enyi maximum correlation coefficient, RDC performs well
empirically, is scalable to very large datasets, and is easy to adapt to concrete
problems. 

\newpage
\clearpage

\appendix
\section{R Source Code}\label{sec:code}
\begin{small}
\begin{verbatim}rdc <- function(x,y,k,s) {
  x  <- cbind(apply(as.matrix(x),2,function(u) ecdf(u)(u)),1)
  y  <- cbind(apply(as.matrix(y),2,function(u) ecdf(u)(u)),1)
  wx <- matrix(rnorm(ncol(x)*k,0,s),ncol(x),k)
  wy <- matrix(rnorm(ncol(y)*k,0,s),ncol(y),k)
  cancor(cbind(cos(x%*%wx),sin(x%*%wx)), cbind(cos(y%*%wy),sin(y%*%wy)))$cor[1]
}
\end{verbatim}
\end{small}

\newpage
\clearpage
\bibliography{nips2013_rdc}

\begin{thebibliography}{10}

\bibitem{Bach02}
F.~R. Bach and M.~I. Jordan.
\newblock Kernel independent component analysis.
\newblock {\em JMLR}, 3:1--48, 2002.

\bibitem{Breiman85}
L.~Breiman and J.~H. Friedman.
\newblock {Estimating Optimal Transformations for Multiple Regression and
  Correlation}.
\newblock {\em Journal of the American Statistical Association},
  80(391):580--598, 1985.

\bibitem{Gebelein41}
H.~Gebelein.
\newblock {Das statistische Problem der Korrelation als Variations- und
  Eigenwertproblem und sein Zusammenhang mit der Ausgleichsrechnung}.
\newblock {\em Zeitschrift f\"ur Angewandte Mathematik und Mechanik},
  21(6):364--379, 1941.

\bibitem{gihman4s:_theor_stoch_proces}
I.I. Gihman and A.V. Skorohod.
\newblock {\em The Theory of Stochastic Processes}, volume~1.
\newblock Springer, 1974s.

\bibitem{Gretton12}
A.~Gretton, K.~M. Borgwardt, M.~J. Rasch, B.~Sch\"{o}lkopf, and A.~Smola.
\newblock A kernel two-sample test.
\newblock {\em JMLR}, 13:723--773, 2012.

\bibitem{Gretton05}
A.~Gretton, O.~Bousquet, A.~Smola, and B.~Sch\"{o}lkopf.
\newblock Measuring statistical dependence with {Hilbert-Schmidt} norms.
\newblock In {\em Proceedings of the 16th international conference on
  Algorithmic Learning Theory}, pages 63--77. Springer-Verlag, 2005.

\bibitem{Haerdle07}
W.~K. H\"{a}rdle and L.~Simar.
\newblock {\em {Applied Multivariate Statistical Analysis}}.
\newblock Springer, 2nd edition, 2007.

\bibitem{Hardoon09}
D.~Hardoon and J.~Shawe-Taylor.
\newblock Convergence analysis of kernel canonical correlation analysis: theory
  and practice.
\newblock {\em Machine Learning}, 74(1):23--38, 2009.

\bibitem{Hastie86}
T.~Hastie and R.~Tibshirani.
\newblock Generalized additive models.
\newblock {\em Statistical Science}, 1:297--310, 1986.

\bibitem{Jones92}
L.~K. Jones.
\newblock {A simple lemma on greedy approximation in Hilbert space and
  convergence rates for projection pursuit regression and neural network
  training}.
\newblock {\em Annals of Statistics}, 20(1):608--613, 1992.

\bibitem{Le13}
Q.~Le, T.~Sarlos, and A.~Smola.
\newblock {Fastfood -- Approximating} kernel expansions in loglinear time.
\newblock In {\em ICML}, 2013.

\bibitem{Mardia79}
K.~V. Mardia, J.~T. Kent, and J.~M. Bibby.
\newblock {\em {Multivariate Analysis}}.
\newblock Academic Press, 1979.

\bibitem{Massart90}
P.~Massart.
\newblock The tight constant in the {Dvoretzky-Kiefer}-wolfowitz inequality.
\newblock {\em The Annals of Probability}, 18(3), 1990.

\bibitem{Nelsen06}
R.~Nelsen.
\newblock {\em An Introduction to Copulas}.
\newblock Springer Series in Statistics, 2nd edition, 2006.

\bibitem{Poczos12}
B.~Poczos, Z.~Ghahramani, and J.~Schneider.
\newblock Copula-based kernel dependency measures.
\newblock In {\em ICML}, 2012.

\bibitem{Rahimi08}
A.~Rahimi and B.~Recht.
\newblock Weighted sums of random kitchen sinks: Replacing minimization with
  randomization in learning.
\newblock {\em NIPS}, 2008.

\bibitem{Renyi59}
A.~R\'enyi.
\newblock On measures of dependence.
\newblock {\em Acta Mathematica Academiae Scientiarum Hungaricae}, 10:441--451,
  1959.

\bibitem{Reshef11}
D.~N. Reshef, Y.~A. Reshef, H.~K. Finucane, S.~R. Grossman, G.~McVean, P.~J.
  Turnbaugh, E.~S. Lander, M.~Mitzenmacher, and P.~C. Sabeti.
\newblock Detecting novel associations in large data sets.
\newblock {\em Science}, 334(6062):1518--1524, 2011.

\bibitem{learningwithkernels}
B.~Sch{\"o}lkopf and A.J. Smola.
\newblock {\em Learning with Kernels}.
\newblock MIT Press, 2002.

\bibitem{Sklar59}
A.~Sklar.
\newblock Fonctions de repartition {\`a} $n$ dimension set leurs marges.
\newblock {\em Publ. Inst. Statis. Univ. Paris}, 8(1):229--231, 1959.

\bibitem{Song12}
L.~Song, A.~Smola, A.~Gretton, J.~Bedo, and K.~Borgwardt.
\newblock Feature selection via dependence maximization.
\newblock {\em JMLR}, 13:1393--1434, June 2012.

\bibitem{stein99:_inter_spatial_data}
M.L. Stein.
\newblock {\em Interpolation of Spatial Data}.
\newblock Springer, 1999.

\bibitem{Szekely10}
G.~J. Sz\'{e}kely and M.~L. Rizzo.
\newblock {Rejoinder: {Brownian} distance covariance}.
\newblock {\em Annals of Applied Statistics}, 3(4):1303--1308, 2009.

\bibitem{Szekely07}
G.~J. Sz\'{e}kely, M.~L. Rizzo, and N.~K. Bakirov.
\newblock Measuring and testing dependence by correlation of distances.
\newblock {\em Annals of Statistics}, 35(6), 2007.

\bibitem{Zhang12}
K.~Zhang, J.~Peters, D.~Janzing, and B.Sch{\"o}lkopf.
\newblock Kernel-based conditional independence test and application in causal
  discovery.
\newblock {\em CoRR}, abs/1202.3775, 2012.

\end{thebibliography}
\bibliographystyle{plain}
\end{document}